%% file: Main.tex
\documentclass[review, 3p]{elsarticle}
\usepackage{tikz}
\usepackage{times}
\usepackage{epsfig}
\usepackage{graphicx}
\usepackage{multirow}
\usepackage{float}
\usepackage{amsmath}
\usepackage{amssymb}
\usepackage{algorithm}

\usetikzlibrary{calc}
\usepackage{algpseudocode}
\usepackage{hyperref}

\journal{Journal of \LaTeX\ Templates}

\begin{document}

\begin{frontmatter}

\title{PointCaps: Raw Point Cloud Processing using Capsule Networks with Euclidean Distance Routing}

\author{Dishanika Denipitiyage$^*$, Vinoj Jayasundara$^{\dagger}$, Ranga Rodrigo$^*$, Chamira U. S. Edussooriya$^{*,\#}$}
\address{$^*$Department of Electronic and Telecommunication Engineering, University of Moratuwa, Sri Lanka \\
$^{\#}$Department of Electrical and Computer Engineering, Florida International University, Miami, FL, USA\\
$^{\dagger}$Department of Computer Science, University of Maryland, College Park, MD, USA}

\begin{abstract}
\input{abstract}
\end{abstract}
\begin{keyword}
Point cloud reconstruction\sep classification\sep capsule networks \sep error routing

\end{keyword}
\end{frontmatter}

%\linenumbers
\section{Introduction}
\input{intro}

\section{Related Work}
\input{related_work}
\section{Method}
\input{Method}
\section{Results and Discussion}
\input{results}

\section{Conclusion}
\input{conclusion}

\bibliographystyle{elsarticle-num}
\bibliography{Mybib}

\end{document}

%% file: abstract.tex
Raw point cloud processing using capsule networks is widely adopted in classification, reconstruction, and segmentation due to its ability to preserve spatial agreement of the input data. However, most of the existing capsule based network approaches are computationally heavy and fail at representing the entire point cloud as a single capsule. We address these limitations in existing capsule network based approaches by proposing PointCaps, a novel convolutional capsule architecture with parameter sharing. Along with PointCaps, we propose a novel Euclidean distance routing algorithm and a class-independent latent representation. The latent representation captures physically interpretable geometric parameters of the point cloud, with dynamic Euclidean routing, PointCaps well-represents the spatial (point-to-part) relationships of points. PointCaps has a significantly lower number of parameters and requires a significantly lower number of FLOPs while achieving better reconstruction with comparable classification and segmentation accuracy for raw point clouds compared to state-of-the-art capsule networks.

%% file: intro.tex
\label{sec:intro}

Point clouds have been widely adopted in computer vision due to their applications in autonomous driving, augmented reality, robotics, and drones. A large variety of 3D sensors (e.g., LiDARs) used in such applications produce raw point clouds as their default output, requiring no additional processing. Even though raw point clouds are disordered and irregular, they are still a popular choice for 3D processing due to their ability to preserve the geometric information in 3D space without any discretization.

Deep learning based raw point cloud processing has gained wide adaptation in object classification, reconstruction, and segmentation. A prominent early attempt at directly processing raw point clouds is Pointnet~\cite{qi2017pointnet}, which learns a spatial representation of a point cloud and aggregates individual features to generate a global representation. One limitation of this work is that it discards the spatial arrangements of the points in local regions while aggregating features through the pooling operation.
However, consideration of spatial arrangements is important because similar local regions have distinct spatial arrangements due to permutation invariance. Following the PointNet~\cite{qi2017pointnet} architecture, PointNet++\cite{qi2017pointnet++} proposed a hierarchical network architecture to combine local features. Furthermore, EdgeConv~\cite{wang2019dynamic} and PointCNN~\cite{li2018pointcnn} proposed a convolutional local feature aggregator based on the neighbourhood graph. Above methods have mainly focused on improving global feature vector via different local feature extracting methods, predominately using k-NN-like clustering techniques to represent the point-to-part relationship. Learnable point-to-part relationship---manifested in the form of capsules~\cite{sabour2017dynamic}---is more powerful in this context. Therefore, we extend the capsule networks~\cite{sabour2017dynamic} to identify spatial relationship in local regions while considering the feature existence of local regions. 

A capsule's ability to learn the spatial relationship of local regions stems from dynamic routing algorithm
which establishes the mapping between the lower level capsules and higher level capsules. In other words, a capsule's activity vector is able to represent a specific type of object or an object part through this routing agreement. 3D-PointCapsNet~\cite{zhao20193d} is the first architecture to formulate capsules with raw point clouds, 
generating the latent representation through fully-connected capsules resembling the multi-layer perceptron architecture. This approach is computationally intensive resulting in longer training and testing time and 3D-PointCapsNet~\cite{zhao20193d} failed to identify proper point to part representation in unordered point clouds.Furthermore, 3D-PointCapsNet~\cite{zhao20193d} fails at representing the entire point cloud as a single capsule causing the latent representation to be not linearly separable and requires a separate SVM for classification. 
A recent work~\cite{wen2020point2spatialcapsule} uses such a representation, but still suffers from high complexity due to feature aggregation through clustering. 
Nevertheless, there are two main problems associated with the capsule networks: 1) Since the logits values in the dynamic routing is bounded, dissimilarity between capsules ranges between $-1$ and $0$. As a result, similarity gap of dissimilar capsules and similar capsules is reduced. 2) The original capsule network implementation assumes static pixel locations, However, point clouds are irregular. 

In order to address these limitations, we propose \emph{PointCaps}: a novel capsule based auto-encoder architecture, which has two novel \emph{convolutional} capsule layers, to capture point-to-part spatial relationships and vice versa. Instead of using a traditional transformation matrix to transform low dimensional features to high dimensional features, our approach adopts the 2D convolution capsule idea into sparse 3D point clouds by creating capsules along the feature axis. 
More importantly, there is a significant reduction in the number of parameters in the capsule layer due to parameter sharing in \emph{convolutional} capsules. This leads to a significant reduction in the \emph{computational complexity} while providing better identification of geometric and spatial relationships between the parts. Furthermore, we employ a novel routing algorithm: dynamic Euclidean distance (\texorpdfstring{$\mathcal{L}_2$}) based routing (ER) in multiple capsule layers instead of dynamic routing (DR) as a solution to the lower similarity gap in dynamic routing. 
This increases the resolution of highly dissimilar capsules in between $-\infty$ and $0$ instead of $-1$ to $0$.
Moreover, we represent the entire point cloud as a single capsule by adopting the approach of Sabour \emph{et al.}~\cite{sabour2017dynamic} and replacing the decoder with a class-independent decoder proposed by Rajasegaran \emph{et al.}~\cite{rajasegaran2019deepcaps}.
PointCaps's ability to compress single point cloud to a vector of instantiation parameters enables us to explore the robustness of the model to noise while completing clasification and reconstruction tasks simultaniously.
To recover lost fine-grained spatial information, we introduce a skip connection between the encoder and the decoder. 
{\bf Our contributions are three-fold:} 
\begin{itemize}
\setlength\itemsep{1em}
\item We propose a novel capsule auto-encoder architecture to classify, reconstruct, and segment raw point clouds. Further, we propose a novel convolution capsule layer with dynamic Euclidean routing instead of dynamic routing to capture part-whole relationships. 
\item To the best of our knowledge, PointCaps is among the first
to adapt a class-independent decoder to reconstruct 3D point clouds. 
\item We evaluate classification accuracy, reconstruction error, and segmentation accuracy (in terms of mean intersection over union (IoU)) of {PointCaps} using standard benchmarks, where our approach surpasses the current state-of-the-art of reconstruction error and provides comparable performance in point cloud classification and segmentation despite having $85\%$ less parameters and requiring $72\%$ less floating point operations per second (FLOPs) compared to previous capsule based architectures. 
\end{itemize}

%% file: related_work.tex
Deep learning applications of point clouds include 3D object detection, object classification~\cite{qi2017pointnet, qi2017pointnet++, wang2021point, lee2021saf}, reconstruction~\cite{chen20153d, qi2017pointnet, cheraghian20193dcapsule}, scene labeling~\cite{lai2014unsupervised, qin2018deep}, segmentation~\cite{qi2017pointnet, qi2017pointnet++, maturana2015voxnet}, point cloud completion~\cite{yuan2018pcn, wu2021point}, layout inference~\cite{geiger2015joint}, and  point cloud registration~\cite{wang2019prnet, zhang2019frequency, maiseli2017recent}. The three main categories of 3D object classification based on the input to the deep learning network are volumetric representation~\cite{maturana2015voxnet, wu20153d},  view-based~\cite{qi2016volumetric} and raw point cloud methods~\cite{qi2017pointnet, wen2020point2spatialcapsule}.
In this paper, we will be focusing on raw 3D point cloud object classification and reconstruction.

\textbf{Deep networks on point clouds:} The capability of processing irregular, unordered point clouds through point-wise convolution and permutation invariant pooling proposed by PointNet~\cite{qi2017pointnet} paved way for various point cloud-speciﬁc architectures such as PointNet++~\cite{qi2017pointnet++}, spherical convolution~\cite{lei2018spherical}, Monte-Carlo convolution~\cite{hermosilla2018monte}, graph convolution~\cite{wang2021point, defferrard2016convolutional} and  SO-Net~\cite{li2018so}. Unlike PointNet, PontNet++~\cite{qi2017pointnet++} aggregate local features into a global feature vector forming a hierarchical feature learning architecture through farthest point sampling. Thereafter, improved convolotion operations~\cite{li2018pointcnn} has been proposed to group local region features. SO-Net~\cite{li2018so} proposed self-organization networks where spatial distribution of point clouds are used in an auto-encoder architecture to enhance the performance. 
A better upsampling method was introduced in PU-Net~\cite{yu2018pu}, and conversion of 2D grid into 3D surface was proposed by FoldingNet~\cite{yang2018foldingnet}. AtlasNet~\cite{groueix2018papier} is an extension of FoldingNet~\cite{yang2018foldingnet} which uses multiple data patches. PPF-FoldNet\cite{deng2018ppf}, which is based on supervised PPFNet~\cite{deng2018ppfnet}, uses FoldingNet decoder~\cite{yang2018foldingnet} to enhance local feature extraction. However, all of the above methods use pooling operation to learn global features from local features based on the feature existence. We focus both existence of features in local regions and their spatial relationship through a capsule-based architecture and employ a new routing algorithm to aggregate the geometric features and spatial relationships in the local region.

\textbf{Capsule networks:} Hinton~\emph{et al.}~\cite{hinton2011transforming} proposed capsule networks, a novel method to group neurons, which greatly impacted object classification in deep learning. Sabour \emph{et al.}~\cite{sabour2017dynamic} extended this idea by proposing dynamic routing between capsules. 
The success of capsule networks in object classification translates well into 3D point cloud classification due to its ability to capture spatial relationships through dynamic routing. Moreover, the instantiating parameters available in these  networks are capable of capturing various properties (e.g., size, position, and texture) of a particular entity. In view of this, our work focuses on a novel auto-encoder architecture to classify and segment raw point clouds, and achieves minimum reconstruction error using capsule networks. 

Several recent works address the use of capsule networks in point cloud classification, reconstruction, and segmentation. 
The 3D-PointCapsNet~\cite{zhao20193d} is the first to devise a capsule network for raw point clouds where part segmentation classification is completed in an unsupervised way. However, it fails to model an entire point cloud as a single capsule. Several previous works~\cite{zhao2019quaternion, cheraghian20193dcapsule} have proposed supervised capsule architectures for point cloud classification. Cheraghian \emph{et al.}~\cite{cheraghian20193dcapsule} applies capsule networks as a drop-in replacement for a fully connected classifier. However, these models are trained in a supervised manner, in contrast to our auto-encoder architecture in PointCaps.  
Point2SpatialCapsule~\cite{wen2020point2spatialcapsule} uses capsule networks to encode fixed spatial locations into capsules. These capsule network architectures\cite{cheraghian20193dcapsule, zhao20193d, wen2020point2spatialcapsule} directly use classification capsule layer with fully-connected capsule architecture for feature representation. This and the presence of k-NN clustering~\cite{wen2020point2spatialcapsule} lead to high computational complexity. On the other hand, convolutional capsule layer (PointCapA and PointCapB described in Sec.~\ref{section:Method}) and the absence of k-NN clustering in PointCaps provides significant reduction in computational complexity.  

%% file: Method.tex
% NEW CONTENT
\label{section:Method}
The 3D point capsule Network \cite{zhao20193d}, has proposed an end to end trainable auto encoder architecture for several common point cloud-related tasks. Inspired by the benefits of capsule network, we propose PointCaps, for processing a point cloud by simultaneous classification and reconstruction, and later achieve segmentation. Point cloud processing differs significantly from regular deep networks based vision tasks due to the irregular and unordered nature of point clouds. The high level processing pipeline of PointCaps: 1) reduces the size of the original data using convolutional capsule networks 2) generates the latent vector representations followed by reconstructing the point cloud using deconvolutions.

In the following sections, we first describe the overall PointCaps architecture. Second, we describe the different types of capsule layers we employ. Finally, we elaborate the routing mechanisms with Euclidean distance. 
\subsection{PointCaps Architecture }
The proposed point-cloud classifier reconstructor network comprises an encoder with Euclidean and dynamic routing, and a class-independent decoder. The encoder architecture contains three types of capsule layers to learn spatial and geometric features of irregular, unordered 3D data. we designed the overall architecture as Fig.~1. The input to PointCaps is a 3D point cloud with $N$ points. We chose $N = 2048$ following the work by Zhao~\emph{et al.}~\cite{zhao20193d}. Two 1D convolution layers process this input to produce a feature vector of length $64$, one for each point. This layer is followed by a \emph{PointCapA} capsule layer: (Fig.~1.: path A) which creates the point-to-part relationships. The second path (Fig.~1.: path B) generates point-to-part relationship through sparse subspace. 
The direct use of 2D capsule layer gives larger parameter space which increases the use of computational resources. Therefore we use PointCapA as a dimension reduction technique in the second path and follow \emph{PointCapC} and \emph{PointCapB} capsule layers to retain the essential properties of parts. Then the parts are regenerated using a PointCapA capsule layer. The two paths: path A and path B are concatenated. Using the concatenated output, the DigitCap capsule layer generates the latent representation of a point cloud. Note that \emph{PointCapB} is an upsampling layer, deployed as an intermediate capsule layer after the \emph{PointCapA} capsule layer, which generates various properties of a given part that is available in the point cloud. Furthermore, \emph{PointCapC} layer is a generic convolutional layer, with squashed output.

It is important to note that the original argument of dynamic routing in capsule network~\cite{sabour2017dynamic} was the capturing of intrinsic geometric properties of the object. These capsule representations sometimes may not correspond to human visible part segments in the object. However, we expect to centralize semantically similar regions in the object so that a human can identify. Furthermore, the coupling coefficient determines the agreement between the current output and the prediction using cosine similarity. As the coupling coefficient and logits are bounded, the gap between highly dissimilar capsules lies within $-1$ and $0$. We increase this gap by making logits unbounded while keeping the coupling coefficient bounded using Euclidean distance. This increases the dissimilarity range between $-\infty$ and $0$. We employ novel \emph{PointCapA}, which predicts the possible point-to-part representation for each point using a given dynamic Euclidean ($\mathcal{L}_2$) distance routing algorithm (see Sec.~\ref{section:Error_Routing}). 

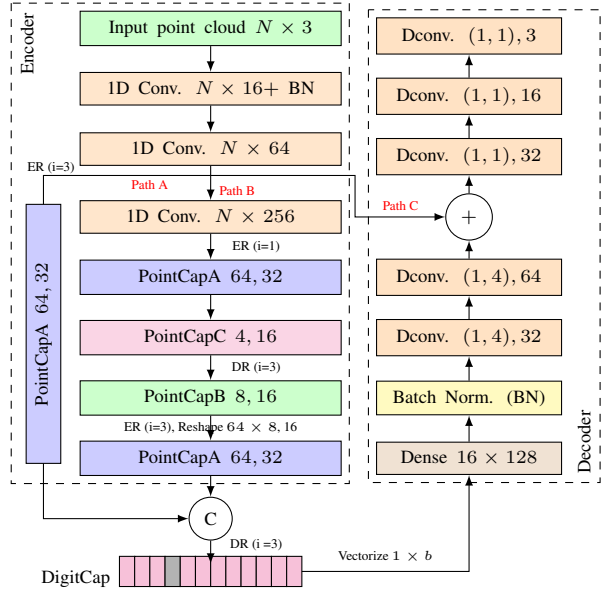
\begin{figure}[t!]
\centering
\label{fig:Point cap architecture}
\input{bd}
\caption{Model Architecture, PointCapA and PointCap B are convolutional capsule layers where PointCapA uses Euclidean routing and in the other capsules dynamic routing is
applied.DR denotes the dynamic routing whereas ER denotes dynamic Euclidean distance routing}
\vspace{-1ex}
\end{figure}

\subsubsection{PointCapA}
\label{sssec:pca}
Let \(\Psi^l\in\mathbb{R}^{c^l\times n^l}\) be the input to the PointCapA layer, where \(c^l\) and \(n^l\) denote the number of input capsules and the input number of atoms (capsule dimension), respectively. Initially, each point with its features is considered as an input to the layer, and \(\Psi^{l+1}\in\mathbb{R}^{c^{l+1}\times n^{l+1}}\) is considered as the output capsule from the layer $l$. The output capsules correspond to different local regions in the point cloud. The activity vector parameter interprets different properties of local region such as size, orientation, and texture. 

The operation of the PointCapA 1D convolutional capsule is as follows. First \(\Psi^l\) is convolved with \((c^{l+1}\times n^{l+1})\) number of \(\psi_l\) kernels, forming   \(\Psi_A^{conv}\in\mathbb{R}^{c^l\times(c^{l+1}\times n^{l+1})}\), where \(\psi_l\in[1,  n^l]\).  Then swish activation function~\cite{ramachandran2017searching} is applied as the pre-activation function to $\Psi_A^{conv}$, and then reshaped to generate the vote matrix \(V_A\) which has the shape of \((c^l,c^{l+1},n^{l+1})\). Using a 1D convolution and choosing the kernel height as one in the transformation matrix has two advantages: 1) it provides a solution to the order invariance problem 2) it allows the network to keep the value of \(c^l\) (input number of capsules) unchanged. Then we feed the vote matrix to the routing algorithm as described in Sec.~\ref{section:Error_Routing}.

Similar to the approach used in \cite{sabour2017dynamic}, the transformation matrix learns the part-to-whole relationships between the lower and higher level capsules by updating the logits based on the similarity between the input capsule and the output capsule.

\subsection{Routing Algorithm}
\label{section:Error_Routing}

\input{ER}

\begin{algorithm}
\caption{Dynamic Euclidean Routing Algorithm}
\label{alg:ER}
\begin{algorithmic}[1]
\Procedure{Routing}{}\\
\algorithmicrequire{ $\mathbf{V}_A\in\mathbb{R}^{c^l\times c^{l+1}\times n^{l+1}}$, $r$ and $l$}
\State $\mathbf{B} \gets \mathbf{0} \in \mathbb{R}^{c^{l+1}\times c^{l}}$
{Let $i \in c^{l}, j \in c^{l+1}$}
\For {$r$ iterations}
\State \text{for all $i$,} $ \  k_{i} \gets \texttt{softmax}(b_{i})$
\State \text{for all }$j,$ $ \ s_{j} \gets \sum_{i} k_{ij} \cdot {v}_{j|i}$
\State \text{for all }$j,$ $ \ \hat{s}_{j} \gets \texttt{squash}(s_{j})$
\State \text{for all }$j,$ $ \ b_{ij} \gets b_{ij} - \ \|{v_A}_{j|i}-\hat{s}_{j}\|_2^2$
\EndFor
\State\Return $\hat{s}_{j}$
\EndProcedure
\end{algorithmic}
\end{algorithm}
% \vspace{-0.4cm}

\subsubsection{PointCapB}
PointCapB is a 2D convolutional capsule layer. We use PointCapB in PointCaps architecture to identify the properties of the entities such as length, elongation and texture. PointCapB operates as follows. Let the input tensor to the PointCapB be  \(\Psi^l\in\mathbb{R}^{E\times c^l\times n^l}\), where \(E\) is the number of entities, \(c^l\) is the number of capsules in the $l^{\mathrm{th}}$ layer and \(n^l\) is the capsule dimension. First, the input tensor \(\Psi^{l}\) is reshaped into \((E, c^l\times n^l, 1)\), where \((E, c^l\times n^l, 1)\) is the standard format of the input for the 2D convolution \((H_{in}, W_{in}, C)\). Then the reshaped tensor is convolved with ($c^{l+1}\times n^{l+1}$) number of \(\psi_i\) 2D kernels, where the size of \(\psi_i\) is \((1\times n^l)\). Note that the height and width of the input feature map for the 2D convolution are represented by $E$ and ($c^l\times n^l$), respectively. Maintaining the kernel height as $1$, width and stride as $n^l$ enable PointCapB to get a vote for single capsule from layer $l$. This process generates intermediate votes \(\Psi_B^{conv}\in\mathbb{R}^{E\times c^l\times (c^{l+1}\times n^{l+1})}\), where the width of the output can be calculated as
\begin{align} \label{output_width}
W_{out} = \frac{c^l\times n^l - n^l+0}{n^l}+1=c^l.
\end{align}
The intermediate votes are then reshaped into votes $V_B$. The vote tensor $V_B$ has the shape of $(1, E, c^l,c^{l+1}, n^{l+1})$. We apply the pre-activation swish function~\cite{ramachandran2017searching}. Then the votes are fed to the routing algorithm as proposed by Rajasegaran~\emph{et al.}~\cite{rajasegaran2019deepcaps}. 

During routing, \emph{softmax} function is applied on logits $B_s \in \mathbb{R}^{1\times E\times c^{l+1}\times c^l}$ for each $s\in c^l$ (initialized as $0$) to generate coupling coefficients $K_s$. Here, we normalize the logits among all the predicted capsules from capsule tensor $S$ in layer $l$. Each generated prediction in $V_B$ is weighted by a factor $k_{prs}\in[0,1]$, which results in a single prediction $S_{pr}$. Then the \emph{squash} function is applied to the single prediction $S_{pr}$. The level of agreement between $S$ and $V_B$ is measured using cosine similarity to update the logits in the next iteration of the routing.

\subsubsection{PointCapC}
\label{section:PointCapC}
Now we describe the architecture of the PointCapC. Let $\Psi^l\in\mathbb{R}^{E\times c^l\times n^l}$ be the input to the PointCapC and $\Psi^{l+1}\in\mathbb{R}^{E\times c^{l+1}\times n^{l+1)}}$ be the output, where $E$ is the number of entities or points, and $c^l$ and $n^l$ have usual meaning. First \(\Psi^l\) is reshaped into a matrix of shape $(E, c^l\times n^l)$, and 1D convolution is applied with $(c^{l+1}\times n^{l+1})$ kernels having the shape $(1,c^l\times n^l)$. Then the output is reshaped into a tensor of shape $(E, c^{l+1}, n^{l+1})$, followed by the \emph{squash} function to produce the output.

\subsection{Class Independent Decoder with Skip Connection}
The decoder network is used to reconstruct the input point cloud using the instantiation vector extracted from the DigitCap in the encoder network. 
In the original capsules~\cite{sabour2017dynamic}, DigitCap is masked to extract activity vectors and then used with three fully-connected layers to reconstruct the input image. 
During the training, they mask the digit capsule output with the true label, and the activity vector of maximum length is used for the inference stage. This vectorization results in a $\mathbb{R}^{a \times b}$ matrix with zeros except for the row corresponding to the true class or predicted class. Here \(a\) is the number of classes and \(b\) is the classification capsule dimension. Hence, the network gets class information which indirectly makes the decoder class dependent. DeepCaps~\cite{rajasegaran2019deepcaps} has claimed that class independent decoder based capsule networks are better for regularization.

To address this issue, in this paper, we use the class independent decoder proposed by Rajasegaran \emph{et al}.~\cite{rajasegaran2019deepcaps} which provides better regularization in terms of capsule encoding. The proposed decoder network uses a class independent network by passing only the activity vector. Here, the masked activity vector is \(P_t\in[1,b]\), where \(t\) is equivalent to true prediction in the training stage whereas \(t=\arg\max_i(\|P_i\|_2^2)\) for testing.
The decoder learns different distributions of different physical parameters irrespective of the class which makes the decoder class-independent. The network consists of a single fully connected network followed by five deconvolution layers. Moreover, the convolution layer from the encoder is skip-connected to the intermediate layer in the decoder as shown in  Fig.~1.
Further, Chamfer distance loss is used as the reconstruction loss and the input point cloud is recreated at the final layer.

{\bf Loss function:} The total auto-encoder loss is defined as the summation of classification and reconstruction losses, i.e., 
\begin{equation} \label{total_loss}
Loss = \sum_{k\in a}L_k+\gamma\,CD(X, Y)
\end{equation}
where $\gamma=0.5$, 
For a each class \(k\) we use margin loss \(L_k\) as indicated in the Eq.~\ref{margin_loss}, where $T_k=1$ if the $k^{th}$ class is present and otherwise zero and $m^+ = 0.9$ and $m^- = 0.1$ are the lower bound and upper bound of the correct and incorrect class. We use $\lambda=0.5$ to reduce the effect of absent classes. The sum of the losses of all digit capsules is defined as total classification loss.
\begin{align} \label{margin_loss}
L_k &= T_k\,\max(0, m^+-\|v_k\|)^2 \notag \\
&\hspace{1cm}+\lambda(1-T_k)\,\max(0, \|v_k\|-m^-)^2
\end{align}
Similar to the other auto encoders, we use Chamfer Distance loss to measure the similarity between point clouds where \(X\) and \(Y\) are two different point clouds with the same number of points. 
\begin{align}\label{chamfer_distance}
CD(X, Y) &= \frac{1}{|X|}\sum_{x\in X} \min_{y\in Y}\|x-y\|_2^2 \notag \\
&\hspace{3mm}+\frac{1}{|Y|}\sum_{y\in Y} \min_{x\in X}\|x-y\|_2^2
\end{align}
\vspace{-0.4cm}

%% file: bd.tex
\begin{tikzpicture}[
	every node/.style={align=center},
	input/.style={rectangle, draw=black, fill=green!20, text width=3.2cm},
	conv/.style={rectangle, draw=black, fill=orange!20, text width=3.2cm},	
	pca/.style={rectangle, draw=black, fill=blue!20, text width=3.2cm},		
	pcb/.style={rectangle, draw=black, fill=green!20, text width=3.2cm},		
	pcc/.style={rectangle, draw=black, fill=magenta!20, text width=3.2cm},	
	concat/.style={circle, draw=black},		
	dense/.style={rectangle, draw=black, fill=brown!23, text width=2.2cm},	
	deconv/.style={rectangle, draw=black, fill=orange!23, text width=2.2cm},
	bne/.style={rectangle, draw=black, fill=yellow!23, text width=3.2cm},
	bnd/.style={rectangle, draw=black, fill=yellow!30, text width=2.2cm},
]
	\def\d{0.8}
	\draw node (input) [input] {\scriptsize Input point cloud $N\times 3$};
	\path (input) ++(0, -\d)	node (conv1) [conv] {\scriptsize 1D Conv. $N\times 16 +$ BN };
% 	\path (conv1) ++(0, -\d)	node (bn1) [bne] {\scriptsize Batch Normalize};
	\path (conv1) ++(0, -\d)	node (conv2) [conv] {\scriptsize 1D Conv. $N\times 64$};	
	\path (conv2) ++(0, -\d-0.1)	node (conv3) [conv] {\scriptsize 1D Conv. $N\times 256$};		
	\path (conv3) ++(0, -\d)	node (pca1) [pca] {\scriptsize PointCapA $64, 32$};		
	\path (pca1) ++(0, -\d)	node (pcc) [pcc] {\scriptsize PointCapC $4, 16$};
	\path (pcc) ++(0, -\d)	node (pcb) [pcb] {\scriptsize PointCapB $8, 16$};	
	\path (pcb) ++(0, -\d)	node (pca2) [pca] {\scriptsize PointCapA $64, 32$};		
	\path (pca2) ++(0, -\d)	node (concat1) [concat] {\scriptsize C};	
	\path (concat1) ++(0, -\d) coordinate (digitcaps) node[anchor=east, xshift=-1.2cm] {\scriptsize DigitCap};	
	\foreach \i in {0,1, ..., 11}
	{
	    \ifthenelse{\i= 3}{
            \draw[fill=black!30] (digitcaps) ++(\i/5 - 6/5, -0.1) rectangle ++(0.2cm, 0.4cm);
        }{
		\draw[fill=magenta!30] (digitcaps) ++(\i/5 - 6/5, -0.1) rectangle ++(0.2cm, 0.4cm);
		}
	}
	\path (pca2) ++(3.4,0.0)	 node (dense) [dense] {\scriptsize Dense $16\times 128$};	
	\path (dense) ++(0, \d)	node (bn2) [bnd] {\scriptsize Batch Norm. (BN)};
	\path (bn2) ++(0,\d)	 node (deconv1) [deconv] {\scriptsize Dconv. $(1,4), 32$};	
	\path (deconv1) ++(0,\d)	 node (deconv2) [deconv] {\scriptsize Dconv. $(1,4), 64$};	
	\path (deconv2) ++(0,\d)	 node (concat2) [concat] {\scriptsize $+$};	
	\path (concat2) ++(0,\d)	 node (deconv3) [deconv] {\scriptsize Dconv. $(1,1), 32$};	
	\path (deconv3) ++(0,\d)	 node (deconv4) [deconv] {\scriptsize Dconv. $(1,1), 16$};	
	\path (deconv4) ++(0,\d)	 node (deconv5) [deconv] {\scriptsize Dconv. $(1,1), 3$};	

	\draw [-latex] (input) -- (conv1);
% 	\draw [-latex] (conv1) -- (bn1);
	\draw [-latex] (conv1) -- (conv2);	
	\draw [-latex] (conv2) -- (conv3)	node [midway, text=red, xshift=0.35cm, yshift=-0.1cm)] {\tiny Path B}node [midway, text=red, xshift=-0.8cm, yshift=-0.02cm)] {\tiny Path A};	
	\draw [-latex] (conv3) -- (pca1) node [midway, xshift=0.6cm, yshift=0cm)] {\tiny ER (i=1)} ;		
	\draw [-latex] (pca1) -- (pcc);			
	\draw [-latex] (pcc) -- (pcb) node [midway, xshift=0.6cm, yshift=0cm)] {\tiny DR (i=3)};			
	\draw [-latex] (pcb) -- (pca2) node [midway, xshift=0cm, yshift=0cm)] {\tiny ER (i=3),    Reshape $64\times 8$, 16};			
	\draw [-latex] (pca2) -- (concat1);	

	\draw [-latex] (concat1) -- ++(0, -0.6) node [midway, xshift=0.6cm, yshift=0.1cm)] {\tiny DR (i =3)};		;
	\draw [-latex]  (digitcaps) ++(6/5, 0.1) -| (dense) node[near start, above] {\tiny Vectorize $1\times b$};
	\draw [-latex] (dense) -- (bn2);
	\draw [-latex] (bn2) -- (deconv1);
	\draw [-latex] (deconv1) -- (deconv2);	
	\draw [-latex] (deconv2) -- (concat2);
	\draw [-latex] (concat2) -- (deconv3);
	\draw [-latex] (deconv3) -- (deconv4);	
	\draw [-latex] (deconv4) -- (deconv5);	

	\draw[-latex] (conv2) ++(0,-\d/2 +0.05) -- ++(1.9,0) |- (concat2) node [midway, text=red, xshift=0.6cm, yshift=0.15cm)] {\tiny Path C};
	\draw[-latex] (conv2) ++(0,-\d/2 +0.05) -| node [midway, xshift=0.1cm, yshift=0.1cm)] {\tiny ER (i=3)} ++(-2.2,-2.1) node (pca3) [pca, rotate=90] {\scriptsize PointCapA $64, 32$} |- (concat1);

	\draw [dashed] ($(input.north west) + (-0.9, 0.1)$)  node [rotate=90, anchor=north east] {\scriptsize Encoder} rectangle ($(pca2.south east) + (0.1, -0.1)$);
	\draw [dashed] ($(deconv5.north west) + (-0.1, 0.1)$)  rectangle ($(dense.south east) + (0.5, -0.1)$) node [rotate=90, anchor=south west] {\scriptsize Decoder} ;

\end{tikzpicture}

%% file: ER.tex
% \label{section:Routing Algorithm}
Routing is a standard method that is used in capsule networks to identify the relevance between a lower level capsule and an upper level capsule~\cite{sabour2017dynamic}. In PointCaps, we employ routing to generate point to parts relationships. Unlike Sabour \emph{et al.}~\cite{sabour2017dynamic}, where the agreement between the current output \(V_j\) and prediction \(v_{A_{j|i}}\) is the dot product between two quantities and the logits are updated based on the measurements for the next iteration, our novel Euclidean distance routing employs the Euclidean distance to find the relevance between capsule layers, and experimentally proving that Euclidean distance provides better performance compared to cosine similarity. 

%describe the motivation of taking these steps why we use these steps
The operation of the routing algorithm is as follows. The routing algorithm maps a block of capsules in the child capsule to the parent capsule. Let the vote tensor (votes) be denoted by $V\in\mathbb{R}^{c^l\times c^{l+1}\times n^{l+1}}$. Following the~\cite{sabour2017dynamic}, we initialize logits $B_s$ as $0$ where \(B\in\mathbb{R}^{c^l\times c^{l+1}}\). Then the corresponding coupling coefficients $K$ are generated by applying the \emph{Softmax} function, defined as 
\begin{equation} \label{softmax}
k_{ij} = \frac{\exp(b_{ij})}{\sum_r \exp(b_{ir})},
\end{equation}
on logits $B$, where \(i\in c^l\) and \(j\in c^{l+1}\)
% As given in Eq. \ref{softmax}, \emph{Softmax} function is used to calculate coupling coefﬁcients \(k_{ij}\in[0,1]\) between capsule \(i\) and all the capsules in the layer above,
This results the iterative dynamic routing process.
Here the logits are normalized over all the predicted capsules in layer $l$ because each single capsule in layer \(l\) predicts the outputs for all the capsules in layer \(l+1\). Then these predictions are weighted by \(k_{ij} \in K\) as 
\begin{equation} \label{preactivate} 
s_j = \sum_i{k_{ij}.v_{j|i}}
\end{equation}
and $s_j$ is applied to the squash function, given by
\begin{equation} \label{squash}
\hat{s}_{ij} = \frac{\|s_j\|^2}{1+\|s_j\|^2}.\frac{s_j}{\|s_j\|}.
\end{equation}
The squash function is a non linear function that ensure the higher probability of the existence of an entity to converge to a length nearly $1$ and lower probabilities t0 get a length of almost $0$. 

Sabour et al.~\cite{sabour2017dynamic} proposed cosine similarity as an agreement between the current output \(V_j\) and the prediction \(v_{A_{j|i}}\). The logits are updated based on this similarity measure. We used Euclidean distance to calculate error between two quantities and updated the logits values using, 

\begin{equation} \label{logits_update}
b_{ij} \leftarrow {b_{ij}-\|{v_A}_{j|i}-\hat{s}_{j}\|_2^2}
\end{equation}

%% file: results.tex
\begin{figure*}[t!]
\minipage{\textwidth}
\raggedright
  \includegraphics[width=\linewidth, scale=0.8]{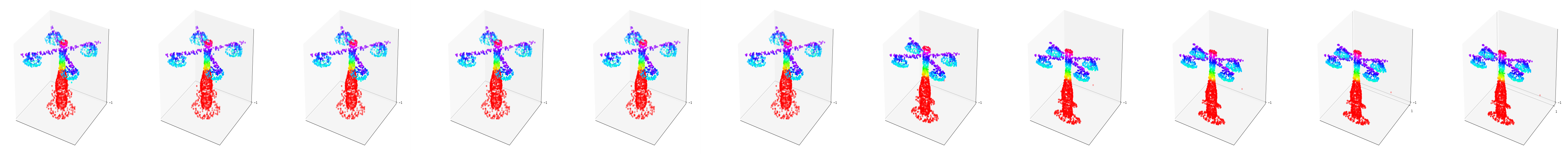}
\endminipage\hfill
\minipage{\textwidth}
\raggedright
  \includegraphics[width=\linewidth, scale=0.8]{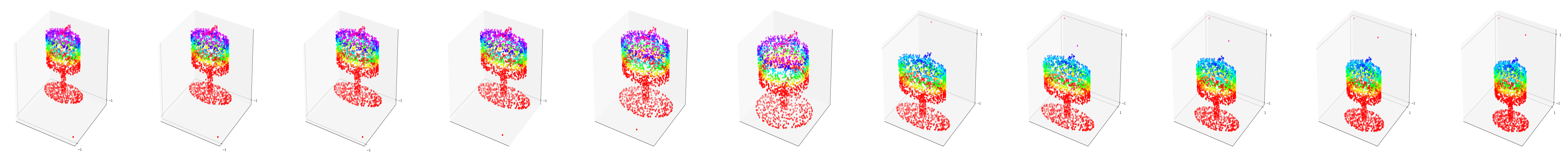}
\endminipage\hfill
% \vspace{-0.4cm}
\caption{Point clouds generated by our decoder. The first row of point clouds is generated using ShapeNet core13~\cite{chang2015shapenet} and the second raw of the point clouds is generated using ShapeNetPart~\cite{yi2016scalable} dataset. This shows the PointCaps ability to capture geometric properties of the point cloud.
For example, we can observe the compression along the $y$-axis when the $26^{th}$ dimension of the instantiation vector of ShapeNetPart lamp is changed between $[-5,5]$.}

\label{fig:generated point clouds}
\vspace*{-0.2cm}
\end{figure*}

\textbf{Implementation: }We implemented our system using Keras and Tensorflow. For training, we used RAdam optimizer~\cite{liu2019variance} with an initial learning rate of $0.001$ and decayed it by $1/10$ over several steps. We conducted an extensive evaluation of our network for point cloud classification and reconstruction using four datasets: ModelNet10~\cite{wu20153d}, ModelNet40~\cite{wu20153d}, ShapeNetPart~\cite{yi2016scalable}, and ShapeNet Core13~\cite{chang2015shapenet}. For ShapeNetPart and ShapeNet Core13, we used $2048$ points, and for ModelNet10 and ModelNet40 we used normal vectors with $1024$ points. The model was trained using two GPUs: Nvidia P100 GPU and Nvidia V100 GPU.

\subsection{Quantitative Evaluations}

\textbf{3D Reconstruction:} We compare our method with both point-cloud based~\cite{deprelle2019learning, zhao20193d, sun2020canonical} and mesh-based~\cite{chen2020bspnet, mescheder2019occupancy, chen2019learning} reconstruction methods using the  Chamfer  Distance matrix. We report quantitative results in Table~\ref{tbl:reconsructionShapeNetCore}. Our method \emph{outperforms the state-of-the-art methods} on the ShapeNet Core13~\cite{chang2015shapenet} dataset. PointCaps ($2048$ points) surpassed the results of AtlasNet~\cite{groueix2018papier} by $79\%$, 3D-PointCapsNet~\cite{zhao20193d} by $83\%$, and Canonical Capsules~\cite{sun2020canonical} by $74\%$. It is worthwhile to note that PointCaps is better than all the mesh-based reconstruction methods and achieves $61\%$ improvement compared to BSPNet~\cite{chen2020bspnet}.

\begin{table}[t!]
\caption{Reconstruction quality of ShapeNet Core13~\cite{chang2015shapenet} dataset. Reconstruction quality is reported by Chamfer Distance (CD) (multiplied by $10^3$). \emph{pc (\#)} denotes point cloud with \# points and \emph{mesh (\#)} denotes mesh with \# vertices}
\begin{center}
\label{tbl:reconsructionShapeNetCore}
\begin{tabular}{lcc}
\cline{1-3}
Method &  Input &  CD \\ \hline
\multicolumn{1}{l}{OccNet~\cite{mescheder2019occupancy}} & mesh (1511) & 2.54 \\
\multicolumn{1}{l}{IM-NET~\cite{chen2019learning}} & mesh (1204) & 2.36 \\
\multicolumn{1}{l}{BSPNet~\cite{chen2020bspnet}} & mesh (1073) & 1.43 \\
\hline
\multicolumn{1}{l}{AtlasNetV2~\cite{deprelle2019learning}} & pc (2048) & 1.22 \\ 
\multicolumn{1}{l}{3D-PointCapsNet~\cite{zhao20193d}}    & pc (2048) & 1.49 \\ 
\multicolumn{1}{l}{Canonical Cap~\cite{sun2020canonical}} & pc (2048) & 0.97 \\ 
\multicolumn{1}{l}{\textbf{PointCaps}} & pc (1024) & 0.56\\
\multicolumn{1}{l}{\textbf{PointCaps}} & pc (2048) & \textbf{0.25}\\ \hline
\end{tabular}
\end{center}
\vspace{-0.5cm}
\end{table}

\textbf{Point Cloud Classification: }We test our model with ModelNet10~\cite{wu20153d} and ModelNet40~\cite{wu20153d} datasets and compare the model with existing network architectures. As shown in Table~\ref{tbl:modelnet10_acc} PointCaps achieves second best performance with \emph{significantly lower computational cost} (See Table~\ref{tbl:pram}) compared to capsule based methods. Moreover, PointCaps is slightly lower than the baseline method PointNet++~\cite{qi2017pointnet++} by $0.2\%$.

\begin{table}[t!]
\caption{ModelNet40 and modelNet10 classification accuracy comparison}
\vspace{-0.1cm}
\begin{center}
\label{tbl:modelnet10_acc}
\begin{tabular}{lcccc}
\hline
Method  & \footnotesize{Input} &  \footnotesize{ModelNet10} & \footnotesize{ModelNet40} \\
\hline
% \hline
PointNet~\cite{qi2017pointnet} & $1024 \times 3$ & - & 89.2\%\\
PointNet++~\cite{qi2017pointnet++}& $1024 \times 3 + n$ & - & 91.9\%\\
DGCNN~\cite{wang2019dynamic} &  $1024 \times 3$  &  - & 92.2\%\\
SAF-Net~\cite{lee2021saf} & $1024 \times 3+n$ & - & 93.4\%\\
Kd-Net~\cite{klokov2017escape} &  $2^{15} \times 3$  &  94.0\% & 91.8\%\\
SO-Net~\cite{li2018so}  &  $2048\times3$  &  94.1\% &  90.9\%\\
PointCNN~\cite{li2018pointcnn} &  $1024 \times 3$  &  - & 91.7\%\\
RS-CNN~\cite{liu2019relation} &  $1024 \times 3$  &  - & 93.6\%\\
Grid-CNN~\cite{xu2020grid} &  $1024 \times 3$   & \textbf{97.5 }\% & 93.1\%\\
CurveNet~\cite{xiang2021walk} &  $1024 \times 3$   & 96.3\% & \textbf{94.2}\%\\
\hline
3DCapsule~\cite{cheraghian20193dcapsule}& $1024\times3$   & 94.7\% &  91.5\%\\
P2SCapsule~\cite{wen2020point2spatialcapsule} & $1024\times3+n$   & \textbf{95.9}\% &  \textbf{93.7}\%\\
\textbf{PointCaps} & $1024\times3+n$ & 94.7\% & 91.7\% \\
\hline
\end{tabular}
% \end{adjustbox}
\end{center}
\vspace{-0.7cm}
\end{table}
 
\begin{table*}[thb!]
\caption{Comparison of classification accuracy and Chamfer Distance (CD) error ($\times 10^3$) of PointCaps with different routing algorithms for three datasets where PointCaps provides better reconstruction and comparable accuracy improvement. Here, \emph{PointCaps} denotes Euclidean routing at PointCapA while dynamic routing (DR) is used in other Caps (see Fig.~1.), 
\emph{All DR} uses DR for all the capsules, and \emph{All ER} uses ER for all the capsules in the model. The performance of \emph{PointCaps} model without using skip connection between encoder and decoder is denoted as \emph{without skip connection}}
\begin{center}
\vspace*{-0.4cm}
% \vspace{-1ex}
\label{tbl:best}

\begin{tabular} {lccccccccc}
\hline
\multicolumn{1}{c}{\multirow{2}{*}{Dataset}} & 
\multicolumn{1}{c}{\multirow{2}{*}{Input}} & 
\multicolumn{2}{c}{PointCaps}                       & \multicolumn{2}{c}{All DR}                    & \multicolumn{2}{c}{All ER}                        & \multicolumn{2}{c}{W/o skip connection}               \\ 
\cline{3-10} 
\multicolumn{1}{l}{} &
\multicolumn{1}{l}{} &
\multicolumn{1}{l}{Accuracy} & \multicolumn{1}{c}{CD} & \multicolumn{1}{l}{Accuracy} & \multicolumn{1}{c}{CD} & \multicolumn{1}{l}{Accuracy} & \multicolumn{1}{c}{CD} & \multicolumn{1}{l}{Accuracy} & \multicolumn{1}{c}{CD} \\ 
\hline
% \hline
ShapeNet core13~\cite{chang2015shapenet} &$2048\times 3$& \textbf{94.12}\% & \textbf{0.25} & \textbf{94.12}\% & 0.34& 93.84\% & 0.315& 94.08\% & 17.42\\
ShapeNet part~\cite{yi2016scalable} &$2048\times 3$ & 98.33\% & \textbf{0.117} & \textbf{98.43}\% & 0.337 & 98.29\% & 0.294 & 98.15\% & 4.07\\
ModelNet10~\cite{wu20153d} &$2048\times 3+n$& \textbf{95.13}\% & 1.71 &  94.59\% & \textbf{1.19} & 94.69\% & 2.11 & 93.90\% & 9.67\\
ModelNet40~\cite{wu20153d} &$2048\times 3$& \textbf{87.6}\% & \textbf{0.891} & 86.52\% & 1.023 & 87.5\% & 0.928 & 86.6\% & 18.13\\
\hline
\end{tabular}
% \end{adjustbox}
\end{center}
\vspace*{-0.4cm}
\end{table*}

\begin{figure*}[t!]
\centering

\begin{tikzpicture}
\node[inner sep=0pt] (pt2part) at (0,0)   {\includegraphics[width=0.8\textwidth]{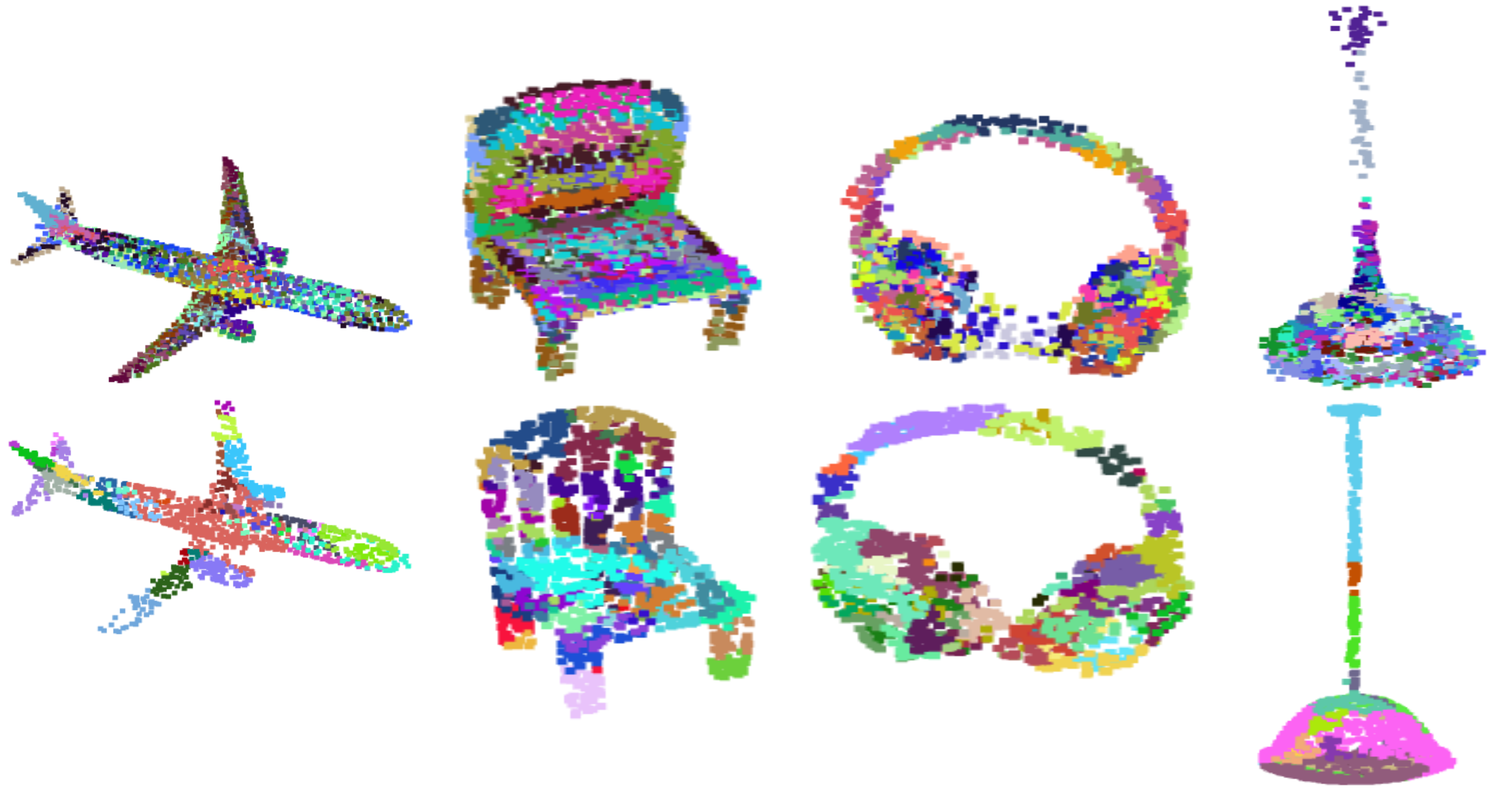}};
\node[fill=white, rotate=90, xshift=0.7cm, yshift=0.5cm] at (pt2part.south west) {PointCap (Ours)};
\node[fill=white, rotate=90, xshift=1.25cm, yshift=0.5cm] at (pt2part.west) {3D-pointCapsNet~\cite{zhao20193d}};
\end{tikzpicture}
\caption{Part representation with dynamic routing in 3D-pointCapsNet~\cite{zhao20193d} (Row 1) and with Euclidean routing in PointCapA (Row 2). 3D-pointCapsNet interprets 32 parts, each having 64 points, whereas PointCaps has 64 parts with different number of points. Note that PointCaps captures spatial relationships to form more human annotated local regions compared to 3D-pointCapsNet~\cite{zhao20193d}}.
\label{fig:point to part}
\vspace*{-0.6cm}
\end{figure*}

\textbf{Segmentation:} In this section we evaluate the part segmentation of PointCaps. We use the  ShapeNet Part dataset to train our model. Following the approach of 3D-PointCapsNet, we train two models: 1) using $1\%$ of dataset (hereafter referred to as $1\%$ training set) and 2) using $5\%$ of dataset (hereafter referred to as $5\%$ training set) as the training set. We used the complete testing dataset set to test our model. We use the same part segmentation evaluation method that is used in SO-Net~\cite{li2018so} and 3D-PointCapsNet~\cite{zhao20193d} to evaluate PointCaps: accuracy and IoU. 
% Table 4 summarizes our results.  
\begin{table}[t!]
\label{tbl:segmentation_results_table}
\begin{center}
\caption{Part segmentation on ShapeNet-Part by learning only on
the $x\%$ of the training data.}
\begin{tabular}{lcccc}
\hline
\multirow{2}{*}{Method} & \multicolumn{2}{c}{1\% data} & \multicolumn{2}{c}{5\% Data} \\ 
\cline{2-5} & Acc & IoU & Acc & IoU \\  \hline
SO-Net~\cite{li2018so} & 0.78 & 0.64 & 0.84 & 0.69 \\ 
3D-PointsCapsNet~\cite{zhao20193d} & \textbf{0.85} & 0.67 & 0.86 & 0.70 \\ 
\textbf{PointCaps} & \textbf{0.85} & \textbf{0.69} & \textbf{0.87} & \textbf{0.72} \\ \hline
\end{tabular}
\end{center}
\vspace{-0.7cm}
\end{table}
As seen in the Table 4, PointCaps surpasses 3D-PointCapsNet~\cite{zhao20193d} and SO-Net~\cite{li2018so} with respect to accuracy and IoU. PointCaps achieves an accuracy of $0.85$ and $0.87$ for the $1\%$ training set and $5\%$ training set, where as the respective values for SO-Net~\cite{li2018so} and 3D-PointCapsNet~\cite{zhao20193d} are $0.78$ and $0.85$ for $1\%$ training set and $0.84$ and $0.86$ for the $5\%$ training set. We also observe that PointCaps achieves better IoU compared to SO-Net~\cite{li2018so} and 3D-PointCapsNet~\cite{zhao20193d}. For the $1\%$ training set, PointCaps achieves an IoU that is $7.81\%$ higher than SO-Net~\cite{li2018so} and $2.98\%$ higher than 3D-PointCapsNet~\cite{zhao20193d}. For the $5\%$ percent training set, Pointcaps surpasses SO-Net~\cite{li2018so} and 3D-PointCapsNet~\cite{zhao20193d} by $4.34\%$ and $2.85\%$ respectively. These results prove that PointCaps achieves better segmentation quality than 3D-PointCapsNet~\cite{zhao20193d} and SO-Net~\cite{li2018so}.

\textbf{Computational Complexity:} 
In this section, we compare the number of model parameters and FLOPs of PointCaps for ModelNet40 classification to recent capsule domain state-of-the-art models. 
Even though Point2SpatialCapsule outperforms in terms of accuracy, PointCaps achieves second best performance whilst \emph{significantly reducing the number of FLOPs}, i.e.,  by $72\%$. 
Moreover, PointCaps has $3.5$ Million parameters that is $85\%$ lower than Point2SpatialCapsule~\cite{wen2020point2spatialcapsule} and $95\%$ lower than 3D-PointCapsNet~\cite{zhao20193d}. 
Table~\ref{tbl:pram} summarizes the results. It is worthwhile to note that, compared to the backbone structure PointNet++, PointCaps has $35\%$ reduction in number of FLOPs. 
Overall, PointCaps achieves performance comparable to the  state-of-the-art models, with significantly lower computational complexity.

\begin{table}[!t]
\caption{The number of model parameters for the ModelNet40 dataset}
\begin{center}
\vspace*{-0.3cm}
\label{tbl:pram}
\begin{tabular}{lcc}
\hline
Method & Params & FLOPs  \\ \hline 
PointNet~\cite{qi2017pointnet}        & 3.48 M & 957 M \\ 
PointNet++~\cite{qi2017pointnet++}      & 1.99 M & 3136 M\\ \hline
3D-PointCapsNet~\cite{zhao20193d} & 69.38 M & 2231 M \\ 
P2SCapsule~\cite{wen2020point2spatialcapsule}      & 22.95 M & 2251 M \\ 
\textbf{PointCaps}       & \textbf{3.52 M} & \textbf{615 M}\\ \hline
\end{tabular}
\end{center}
\vspace{-4ex}
\end{table}

\begin{figure}[]
\centering
  \includegraphics[height=3.8cm, width=\linewidth]{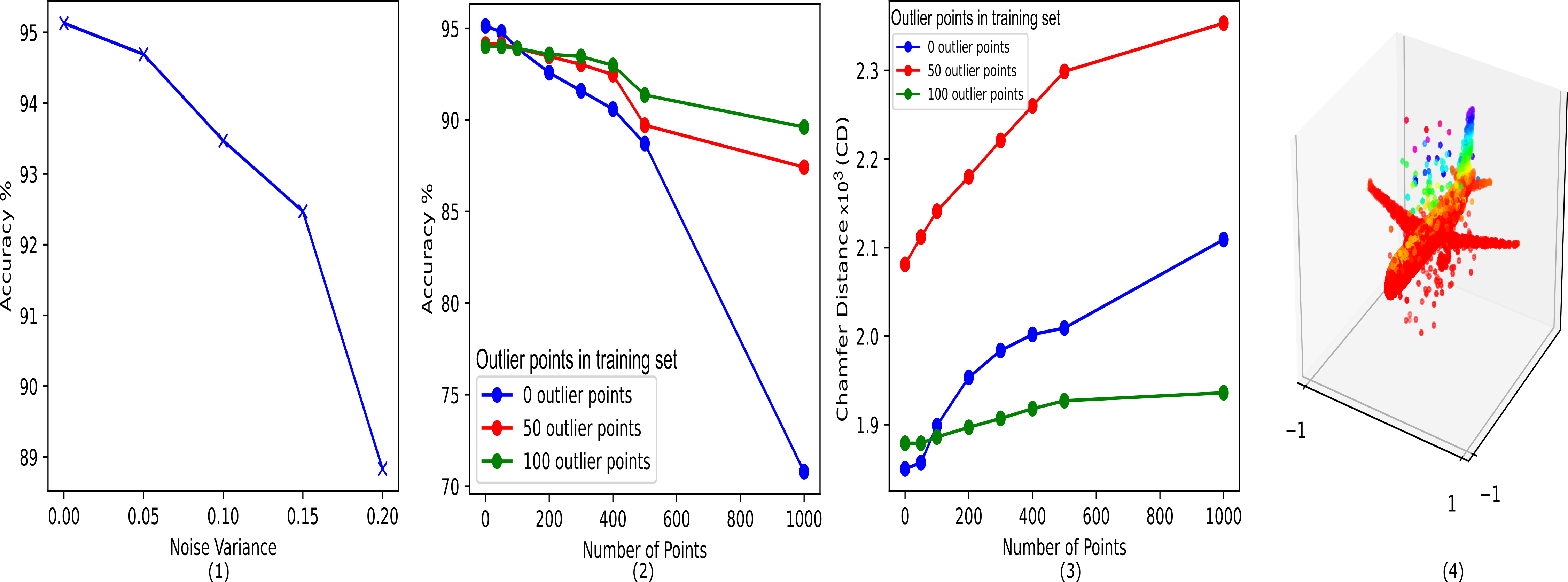}
\caption{Noise analysis on ModelNet10 dataset. (1) The network is trained without any perturbation and tested with Gaussian noise, with variance in the range $0-0.2$.  (2-3) The network is subjected to different number of anomaly points (in the X axis we increase the number of outlier points in the test set) and the performance (accuracy and reconstruction) is analyzed on ModelNet10 dataset. The network is trained with various number of outliers with Gaussian noise $\mathcal{N}(0, 0.2)$. (4) An example of 100 Points replaced with Gaussian noise $\mathcal{N}(0, 0.2)$.}
\label{fig:noise analysis}
\vspace{-0.5cm}
\end{figure}

\textbf{Robustness to Noise: }To evaluate the robustness of our architecture to noise, we train a noise-free version of ModelNet10~\cite{wu20153d} dataset using two augmentation techniques; 1) point perturbation and 2) adding outliers, and evaluate the reconstruction loss and the accuracy matrix. In the perturbation test, Gaussian noise $\mathcal{N} (0, \sigma)$ is added to the points where \(\sigma \in [0, 0.2]\). As shown in Fig.~\ref{fig:noise analysis}.(1), even though the network shows a considerable accuracy drop when $\sigma >= 0.15$, the network still achieves a minimum of  $89.1\%$ accuracy.  
Our outlier test replaces various numbers of points in both training and testing sets. Fig.~\ref{fig:noise analysis}.~(2,3) depicts this behaviour. 
In Fig.~\ref{fig:noise analysis}.~(2, 3) the $X-$axis denotes the number of outlier points in the test set. The three colours represent different number of outliers in the training set. As shown in Fig.~\ref{fig:noise analysis}.~(2, 3),
PointCaps delivers more than $90\%$ accuracy up to $400$ outliers in the test set.
We also observe that the accuracy increases when we add outliers during the training phase. Hence, we conclude that the PointCaps is significantly more robust to Gaussian noise and to anomalies and provides good reconstruction.
% We conclude that the proposed network is significantly more robust to Gaussian noise and to anomalies and provides better reconstruction.

\textbf{Data Generation by Perturbation:} We analyze the ability of PointCaps to generate data by perturbing the instantiation parameters. To experiment on that, we add random noise to only one non-zero instantiation parameter at a time. 
As seen in Fig.~\ref{fig:generated point clouds}, we can observe that the instantiation parameter creates specific changes in the reconstructed point cloud. Furthermore, we observe that the new data samples are not distorted. This proves that latent representation of the PointCaps is capable of capturing interpretable geometric properties and the PointCaps augmenting data with less distortion. We achieve low distortion in data augmentation by applying an upper bound of noise to the instantiation parameter where the maximum variance of noise was manually inspected.

\textbf{Points to Part Capsule:} Here we analyse the capability of PointCapA (Path A in Fig.~1) at representing point-to-part relationship with dynamic Euclidean routing. We compare the ability to represent point to part relationships of PointCaps (Euclidean routing based) with 3D-PointCapsNet (dynamic routing based). 
The 3D-PointCapsNet generates symmetric local regions. This is due to the fact that they only consider existance of geomatrical information and disregard spatial relationships. This problem rises when pooling based methods are used to agregate features in local regions. It filters out features of different areas which represent the existence of characteristics in local regions while ignoring spatial relationships among local regions. However, in PointCaps capable of identifing different spatial arrangements in geometrically similar local regions through dynamic ER routing. The PointCapA is responsible for representing points-to-part relationships.
Fig.~\ref{fig:point to part} illustrates the local part representation of capsules. As indicated in the Sec. 3.2 of the paper, each parent capsule has a logit which increases for the possible parent during routing. This represents the contribution of the lower level capsules to the higher level capsule. We use this logits to identify the relevant part labels for each point. As shown in Fig.~\ref{fig:point to part}, PointCapA is capable of specializing on the local regions compared to 3D-PointCapsNet (dynamic routing based).

\subsection{Ablation Study}

We first evaluate the impact of our novel routing algorithm, Euclidean distance routing (ER) for the accuracy and CD error. To evaluate this, we compare the accuracy and CD error of three implementations; 1) \emph{PointCaps}: where PointCapA uses ER while PointCapB and Digitcap use DR, 2) \emph{All DR}: where all capsule layers use DR and 3) \emph{All ER}: where all capsule layers use ER. As shown in Table~\ref{tbl:best}, the accuracy of PointCaps is slightly above or on-par compared to two other routing techniques. More notably, PointsCaps considerably surpasses All DR network (both accuracy and Chamfer distance) for the benchmark dataset, ModelNet40. Further, except for ModelNet10 data set, PointCaps achieves the best CD. These observations confirm that the use of ER for PointCaps achieves better accuracy and CD. Moreover, we observe that PointCaps provides faster convergence for all the datasets. 

Secondly, we evaluate the impact of skip connection on accuracy and CD error. We compare two implementations; 1) \emph{PointCaps}: where a skip connection is used between the encoder and decoder and 2) \emph{W/o-skip-connection}: which does not contain a skip connection. In Table~\ref{tbl:best}, we show that PointCaps concurrently achieves better CD error ($80\%$ improvement) for all the datasets. This observation proves our previous intuition for using a skip connection; the use of a skip connection results in lower reconstruction error.

%% file: conclusion.tex
In this work, we presented a novel capsule network based architecture for raw point cloud reconstruction, classification, and segmentation. Our approach of using 1D convolutional capsule architecture helps to significantly reduce computational complexity while retaining the global context. Our PointCapsA layers is capable of representing human-interpretable point-to-part relationships. We also introduced a novel routing mechanism, dynamic Euclidean distance routing (as opposed to dynamic routing), and class-independent latent representation. These improved reconstruction, classification, and segmentation accuracy of raw point clouds. Further, our proposed architecture is capable of augmenting data by perturbing instantiation parameters with no distortion.